\documentclass[letterpaper]{article} % DO NOT CHANGE THIS
\usepackage{aaai2026}  % DO NOT CHANGE THIS
\usepackage{times}  % DO NOT CHANGE THIS
\usepackage{helvet}  % DO NOT CHANGE THIS
\usepackage{courier}  % DO NOT CHANGE THIS
\usepackage[hyphens]{url}  % DO NOT CHANGE THIS
\usepackage{graphicx} % DO NOT CHANGE THIS
\usepackage{natbib}  % DO NOT CHANGE THIS AND DO NOT ADD ANY OPTIONS TO IT
\usepackage{caption} % DO NOT CHANGE THIS AND DO NOT ADD ANY OPTIONS TO IT
\usepackage{algorithm}
\usepackage{algorithmic}

\usepackage{newfloat}
\usepackage{listings}
\DeclareCaptionStyle{ruled}{labelfont=normalfont,labelsep=colon,strut=off} % DO NOT CHANGE THIS
\floatstyle{ruled}
\newfloat{listing}{tb}{lst}{}
\floatname{listing}{Listing}
\usepackage{amssymb}
\usepackage{amsmath}
\usepackage{booktabs}
\usepackage{bbding}
\usepackage{multirow} 
\usepackage{subcaption}

\title{DensiCrafter:~Physically-Constrained Generation and Fabrication of Self-Supporting Hollow Structures}
\author {
    Shengqi Dang\textsuperscript{\rm 1,\rm 2,\rm 3},
    Fu Chai\textsuperscript{\rm 2,\rm 3},
    Jiaxin Li\textsuperscript{\rm 2},
    Chao Yuan\textsuperscript{\rm 1,\rm 3},
    Wei Ye\textsuperscript{\rm 1,\rm 2,\rm 3},
    Nan Cao\textsuperscript{\rm 1,\rm 2,\rm 3}\thanks{Nan Cao is the corresponding author.}
}
\affiliations {
    \textsuperscript{\rm 1} Shanghai Research Institute for Intelligent Autonomous Systems\\
\textsuperscript{\rm 2} Shanghai Innovation Institute\\ \textsuperscript{\rm 3} Tongji University\\
    dangsq123@tongji.edu.cn,  
    nan.cao@gmail.com
}

\usepackage{bibentry}
\begin{document}

\maketitle

\begin{abstract}
The rise of 3D generative models has enabled automatic 3D geometry and texture synthesis from multimodal inputs~(e.g., text or images). However, these methods often ignore physical constraints and manufacturability considerations. In this work, we address the challenge of producing 3D designs that are both lightweight and self-supporting. We present DensiCrafter, a framework for generating lightweight, self-supporting 3D hollow structures by optimizing the density field. Starting from coarse voxel grids produced by Trellis, we interpret these as continuous density fields to optimize and introduce three differentiable, physically constrained, and simulation-free loss terms. Additionally, a mass regularization penalizes unnecessary material, while a restricted optimization domain preserves the outer surface. Our method seamlessly integrates with pretrained Trellis-based models~(e.g., Trellis, DSO) without any architectural changes. In extensive evaluations, we achieve up to 43\% reduction in material mass on the text-to-3D task. Compared to state-of-the-art baselines, our method could improve the stability and maintain high geometric fidelity. Real-world 3D-printing experiments confirm that our hollow designs can be reliably fabricated and could be self-supporting.
\end{abstract}

\begin{links}
    % \link{Code}{https://github.com/dangsq/DensiCrafter}
    \link{Code}{ https://github.com/idvxlab/DensiCrafter}
    % \link{Extended version}
    % \link{Extended version}{https://arxiv.org/abs/2511.09298}
\end{links}

\section{Introduction}
Recent advances in 3D generative models, such as Trellis~\cite{trellis} and CLAY~\cite{clay}, have enabled the automatic synthesis of high-quality surface geometry and textures from diverse inputs (e.g., text or images). These capabilities dramatically lower the barrier for designers and creators to prototype rich 3D assets, supporting applications in virtual reality, gaming, and digital art. In addition to generating virtual assets, such 3D generative models can also serve as the front-end for downstream fabrication workflows, such as 3D printing and robotic assembly.

However, bringing these generated assets into the physical world demands more than visual plausibility. It requires adherence to real‐world physics, including material properties, structural mechanics, load‐bearing behavior, and overall manufacturability, which current pretrained 3D generative models seldom consider. Recent works begin to close this gap. Static stability under gravity (i.e., self‐supporting) is often the first constraint imposed: Atlas3D~\cite{atlas3d} uses simulation‐based optimization to progressively stabilize generated meshes, while DSO~\cite{dso} further finetunes the Trellis to generate more stable structures. However, previous methods assume the generated 3D assets as solid interiors and only adjust surface meshes. In reality, the mass distribution is equally critical to physical performance. Moreover, efficient material usage remains essential for fabrication and sustainability. 
However, to the best of our knowledge, there are no solutions that integrate density‐based optimization directly into 3D generative models.

From the perspective of manufacturability and physical constraints, we raise the question: \emph{can we automatically generate lightweight, self‐supporting structures while preserving the high‐quality outer surface produced by modern 3D generators?} Even when restricted to rigid bodies, achieving this is highly non-trivial and presents several key challenges:
(1) Most 3D generative models focus solely on surface geometry, making it difficult to incorporate internal mass distribution into their pipelines.
(2) Enforcing self-supporting behavior typically requires rigid-body dynamics simulations, which are computationally expensive and poorly suited to continuous mass distribution.
(3) A careful balance must be struck between optimizing internal mass and preserving the original surface geometry. Modifying the mass distribution should not compromise the visual fidelity of the generated mesh, which is not an easy task.

To address these challenges, we present \emph{DensiCrafter}, a method for generating lightweight, self-supporting hollow structures from multimodal inputs. Our approach builds on the pretrained state-of-the-art 3D generative model Trellis~\cite{trellis}, which employs a two‐stage pipeline: (1) synthesizing a voxel grid to capture object occupancy, (2) reconstructing the high‐resolution mesh and textures based on the voxel grid. Our method represents Trellis's voxel grid output as a continuous density field and performs online optimization to incorporate physically-aware internal structure into the generative process. To promote self-supporting behavior without resorting to expensive simulations, we introduce three differentiable, simulation-free constraints: (1) shifting the center of mass into the support region, (2) maximizing the bottom contact area, and (3) minimizing the vertical position of the mass center. To ensure material efficiency while preserving visual quality, we penalize total mass and constrain optimization to the object's interior and a thin bottom region, preserving the outer surface. The optimized density field is converted back into a voxel grid and fed into Trellis to reconstruct the high-resolution outer mesh, while the internal hollow geometry is extracted from the optimized density. Our method achieves up to a 43\% reduction in material mass and improves upright stability over all baselines, with minimal impact on inference time. The generated structures retain high semantic alignment and geometric fidelity. Real-world FDM 3D printing further confirms the reliability and manufacturability of the resulting designs (Fig.~\ref{fig:enter-label},~\ref{fig:rotation}). The main contributions are :

\begin{itemize}
\item We introduce a novel task of generating lightweight, self-supporting 3D structures, and propose DensiCrafter, a framework that integrates density field optimization seamlessly into the existing 3D generation pipeline.

\item We design a set of fully differentiable, simulation-free, and physically-constrained losses that guide the optimization by restricting the center of mass and the bottom contact surface, ensuring static stability.

\item We incorporate two regularization terms, including a mass penalty and a spatially restricted optimization domain, achieving material efficiency with surface fidelity.

\item We validate our method through extensive evaluations, and real-world FDM 3D printing experiments demonstrate substantial reductions in material mass, improved upright stability, and practical manufacturability.
\end{itemize}

\begin{figure*}[tbp]
    \centering
    \includegraphics[width=1.0\linewidth]{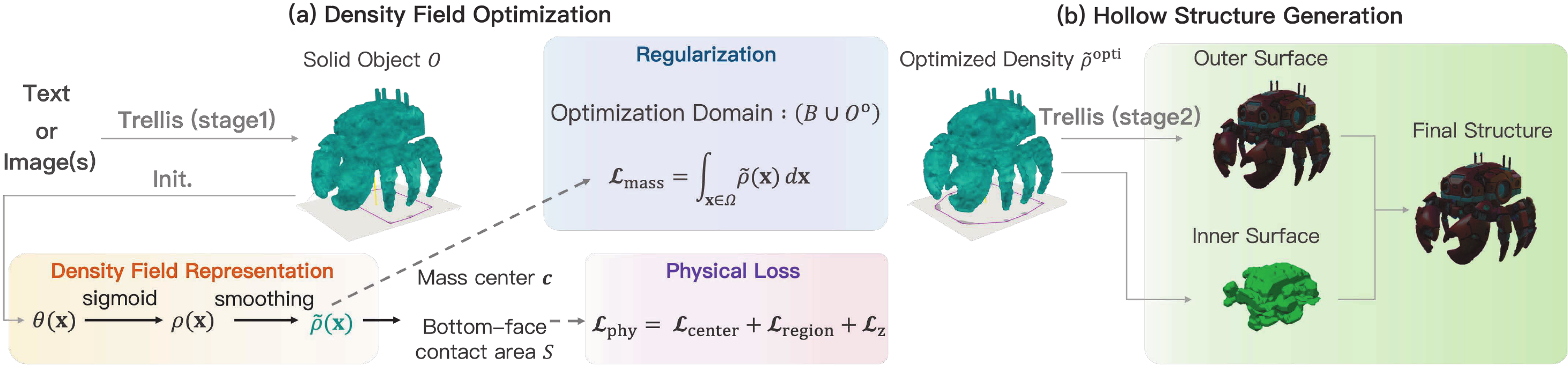}
    \caption{Pipeline of our method.   
    We optimize this density field using differentiable losses that embed simulation-free physical constraints, yielding a material-efficient, self-supporting result.  
}
    \label{fig:pipeline}
\end{figure*}

\section{Related Work}
We review the related works from the following two aspects.
\subsection{3D Generative Model}

Current 3D generative models~\cite{trellis,clay,hunyuan3d} can produce high-quality 3D representations, such as NeRF~\cite{nerf}, 3DGS~\cite{3dgs}, CAD instructions~\cite{deepcad}, or Tetrahedron~\cite{tetsphere}, conditioned on various inputs. Due to the distinct nature of geometry and texture, researchers often decouple their generation processes~\cite{clay,hunyuan3d}. Trellis~\cite{trellis} introduces a structured latent variable approach, dividing generation into two stages: coarse voxel generation followed by fine-grained geometry and material synthesis. Leveraging Trellis's strong generative performance, we can optimize the coarse voxel structure to adjust density for desired physical properties, and then generate textures and detailed geometry conditioned on the refined voxel structure.

\subsection{Physics-aware Generation}
Recently, there has been growing interest in endowing 3D generative models with physically plausible behavior. Some works directly infer an object’s material properties and geometry from input images or video~\cite{image2mass,spring-mass,physical,Pac-nerf}, or reconstruct a complete 3D shape and then subject it to physics-based post-processing under scene constraints~\cite{Physgaussian,Pie-nerf}. Others first optimize existing 3D representations to synthesize new training datasets~\cite{Physics-Based-Validation}, and train a generative model that inherently respects physical requirements~\cite{Physics-Aware}. Depending on the application, different physical attributes are targeted. For example, Atlas3D~\cite{atlas3d}, PhysComb~\cite{physcomp}, and DSO~\cite{dso} focus on ensuring that generated shapes remain statically stable under gravity. However, these methods typically adjust only the surface geometry of the 3D object, assuming the object is a solid of uniform density.

We enhance Trellis's workflow by integrating topology optimization concepts~\cite{TO1,sheparddf,TO2,musialski2016non,bacher2014spin,hafner2024spin}. The traditional topology optimization requires explicit problem specification—including boundary conditions, loads, and mechanical constraints—to optimize material distribution, and is typically applied as a post-processing step after shape design. In contrast, our method incorporates topology-aware considerations directly into the content generation process, without requiring predefined mechanical constraints.

\section{Methodology}
As shown in Fig.~\ref{fig:pipeline}, DensiCrafter builds a lightweight, self‐supporting 3D model for fabrication via two key steps:  
\emph{(a) Density Field Optimization:} We first convert the solid object \(O\) (a set of points within a \(N^3\) voxelized cubic domain \(\Omega\)) generated by Trellis into a continuous density field \(\tilde{\rho}\). This field is then optimized to enhance upright stability and reduce material usage, subject to physical constraints, while preserving the original surface geometry. \emph{(b) Hollow Structure Generation:} 
Given the optimized density field \(\tilde{\rho}^{\mathrm{opti}}\), we extract a high-resolution outer surface mesh along with a complementary inner surface mesh, resulting in a lightweight, self-supporting structure ready for fabrication. The following sections detail each of these steps.

\subsection{Density Field Optimization}
Our method casts a 3D shape as a continuous density field and optimizes the density field by solving
\begin{equation}
\min_{\theta(\mathbf{x})\,:\,\mathbf{x}\in \Omega^{\mathrm{opti}}}
\;\mathcal{L}_{\mathrm{phy}}
+\lambda_{\mathrm{mass}}\,\mathcal{L}_{\mathrm{mass}}    
\end{equation}
where \(\theta(\mathbf{x})\) is the field of parameters that defines the density field $\tilde{\rho}$ at each spatial point \(\mathbf{x}\). The term \(\mathcal{L}_{\mathrm{phy}}\) is the physical loss to enforce self-supporting behavior and upright stability. The term \(\mathcal{L}_{\mathrm{mass}} \) is a loss to minimize the amount of materials used for fabrication, weighted by \(\lambda_{\mathrm{mass}}\). The optimization domain \(\Omega^{\mathrm{opti}} \) restricts updates to the object’s interior and a thin basal layer, preserving the geometry of the outer surface. 
Next, we first introduce the use of $\theta(\mathbf{x})$ to represent the continuous density field. We then detail the physical loss \(\mathcal{L}_{\mathrm{phy}}\) used to enforce self-supporting behavior. Finally, we describe two regularization strategies: the mass penalty term \(\mathcal{L}_{\mathrm{mass}} \) and the restriction of the optimization domain $O^{\rm{opti}}$.

\paragraph{{Density Field Representation.}}
The mass distribution is represented within the cubic domain $\Omega$, as a continuous density field $\tilde{\rho}: \mathbf{x}\to [0,1]$, where $\tilde{\rho}(\mathbf{x}) \approx 1$ indicates solid material and $\tilde{\rho}(\mathbf{x}) \approx 0$ denotes empty space. This density field can be parameterized by:
\begin{equation}
    \rho(\mathbf{x}) = \rm{sigmoid}(\theta(\mathbf{x})) = \frac{1}{1+\exp(-\theta(\mathbf{x}))}
\end{equation}
where $\theta: \mathbf{x}\to  \mathbb{R}$ is the scale field. We initialize $\theta$ to preserve the surface structure of $O$:
\begin{equation}
    \theta(\mathbf{x}) \overset{\rm{init.}}{=}
\begin{cases}
+\infty & \mathbf{x}\in \partial O\\
-\infty & \mathbf{x}\in \Omega\setminus\partial O
\end{cases}
\end{equation}
where the point set $\partial O$ denotes the boundary of $O$.
To promote spatial coherence and avoid isolated material regions, we apply a local smoothing operator to $\rho$, resulting in $\tilde{\rho}$ that is used for defining our loss functions:
\begin{equation}
\tilde{\rho}(\mathbf{x}) = 
\begin{cases}
\rho(\mathbf{x}) & \mathbf{x}\in \partial O\\
\bigl(K * \rho\bigr)(\mathbf{x}), & \mathbf{x}\in \Omega^{\rm{opti}}\\
0 & \rm{otherwise}
\end{cases}
\label{eq:1}
\end{equation}
In our implementation, the smoothing operator applies a $3\times3\times3$ averaging kernel $K$ to $\Omega^{\rm{opti}}$, while preserving the object's original boundary.

\paragraph{\textbf{Physical Loss.}} 
The physical loss 
\(\mathcal{L}_{\mathrm{phy}}\) comprises three differentiable, simulation‐free terms that enable the self-supporting behavior: (1) aligning the center of mass $\mathbf{c}$ within the support region, i.e., the center loss (\(\mathcal{L}_{\mathrm{center}}\)), (2) maximizing the bottom contact region $S$, i.e., the region loss (\(\mathcal{L}_{\mathrm{region}}\)), and (3) lowering the height of the center of mass, i.e., the height loss (\(\mathcal{L}_{z}\)), which is formally defined as:
\begin{equation}
\mathcal{L}_{\mathrm{phy}}
= \mathcal{L}_{\mathrm{center}}
+ \mathcal{L}_{\mathrm{region}}
+ \mathcal{L}_{z}
\end{equation}
Here, the bottom contact region $S$ and the mass center $\mathbf{c}$ can be calculated as:
\begin{equation}
    S = \bigl\{\mathbf{s}\in\Omega : s_z = z_{\min},\ \tilde{\rho}(\mathbf{s}) > 0.5 \bigr\}
\end{equation}
\begin{equation}
    \mathbf{c} = \frac{\displaystyle\int_{\Omega} \mathbf{x}\,\tilde{\rho}(\mathbf{x})\,\mathrm{d}V}
{\displaystyle\int_{\Omega} \tilde{\rho}(\mathbf{x})\,\mathrm{d}V}
\end{equation}
where \(z_{\min}\) means the ground‐level of \(O\) and $\mathrm{d}V$ denotes the differential volume element.

\textit{{Centering Loss.}}
Directly constraining the center of mass to lie inside the support region~(i.e., convex hull of $S$) is not easily differentiable. Instead, we observe that the centroid of $S$ always lies within the support region~(proof provided in the appendix). We penalize the projection of $\mathbf{c}$ onto the bottom plane close to the centroid of $S$:
\begin{equation}
\mathcal{L}_{\mathrm{center}}
= \frac{1}{|S|}\int_{\mathbf{s}\in S}
\bigl\lVert (s_x,s_y) - (c_x,c_y)\bigr\rVert_2
\,\mathrm{d}A
\end{equation}
where $\mathrm{d}A$ denotes the differential area element.

\textit{{Region Loss.}}
A larger support region increases the object's ability to withstand larger perturbations. Since for any two sets $S_1 \subset S_2$ we have $\mathrm{ConvexHull}(S_1)\subset \mathrm{ConvexHull}(S_2)$, expanding the bottom-face contact area directly enlarges the support region. We thus encourage a large contact area by minimizing its negative:
\begin{equation}
\mathcal{L}_{\mathrm{region}}
= -\,|S|
= - \int_{\mathbf{s}\in S} \mathrm{d}A
\end{equation}

\textit{{Height Loss.}}
Lowering the center of mass increases the critical overturning angle, thereby enhancing stability against rotational perturbations. Therefore, we penalize $c_z$:
\begin{equation}
\mathcal{L}_{z}
= c_z
\end{equation}

\paragraph{\textbf{Regularization.}}
To minimize the total mass while preserving the overall shape, we introduce two regularizations: the mass penalty and the restricted optimization domain.

\textit{{Mass Penalty.}}
We directly penalize the total mass of the density field via the loss term $\mathcal{L}_{\mathrm{mass}}$:
\begin{equation}
\mathcal{L}_{\mathrm{mass}}
= \int_{\Omega} \tilde{\rho}(\mathbf{x}) \,\mathrm{d}V
\end{equation}

\textit{{Restricted Optimization Domain.}}
Instead of optimizing the entire voxel grid, we restrict updates to the object’s interior \(O^\circ\) and a thin basal neighborhood \(B\), thereby preserving the external surface structure. The optimization domain $\Omega^{\text{opti}}$ is defined as  
\begin{equation}
    \Omega^{\mathrm{opti}} = (O^\circ \;\cup\; B)
\end{equation}
where  $O^\circ = O \setminus \partial O$
is the interior of the original solid object $O$, and  
\begin{equation}
\begin{aligned}
B = \bigl\{ \mathbf{x}&\in\Omega\setminus O :
z \in [z_{\min},z_{\min} + \epsilon], \\
&\exists\,\mathbf{x}'\in O, s.t. ||\mathbf{x}-\mathbf{x}'||_1 < \epsilon\bigr\}
\end{aligned}
\end{equation}
 is a thin layer above the ground plane that lies beneath existing material, essential for the supporting structure.

\subsection{Hollow Structure Generation}

After optimization has converged, we generate the final hollow structure in three stages.
(1) We compute the solid and hollow regions, based on the optimized density field $\tilde{\rho}$. Specifically, the solid region is defined as $V = \{\mathbf{x} \in \Omega \mid \tilde{\rho}(\mathbf{x}) > 0.5\}$, and the hollow region as $H = \{\mathbf{x} \in O \mid \tilde{\rho}(\mathbf{x}) \le 0.5\}$.
(2) We extract the outer surface mesh $\mathcal{M}_{\mathrm{outer}}$ from $V$ using Trellis~\cite{trellis}, and the inner surface mesh $\mathcal{M}_{\mathrm{inner}}$ from $H$ using Marching Cubes~\cite{marchingcubes}.
(3) We invert the normals of $\mathcal{M}_{\mathrm{inner}}$ and combine it with $\mathcal{M}_{\mathrm{outer}}$ to construct the final structure.
Since the Trellis-generated outer surface aligns precisely with the voxels, the final outer surface resides strictly within the voxel cells corresponding to \(\partial O\). This precise alignment facilitates non‐intersection between the inner and outer surfaces. In our experiments, we did not observe any intersections or perforations between the inner and outer surfaces.

\begin{figure}[t]
    \centering
\includegraphics[width=1.0\linewidth]{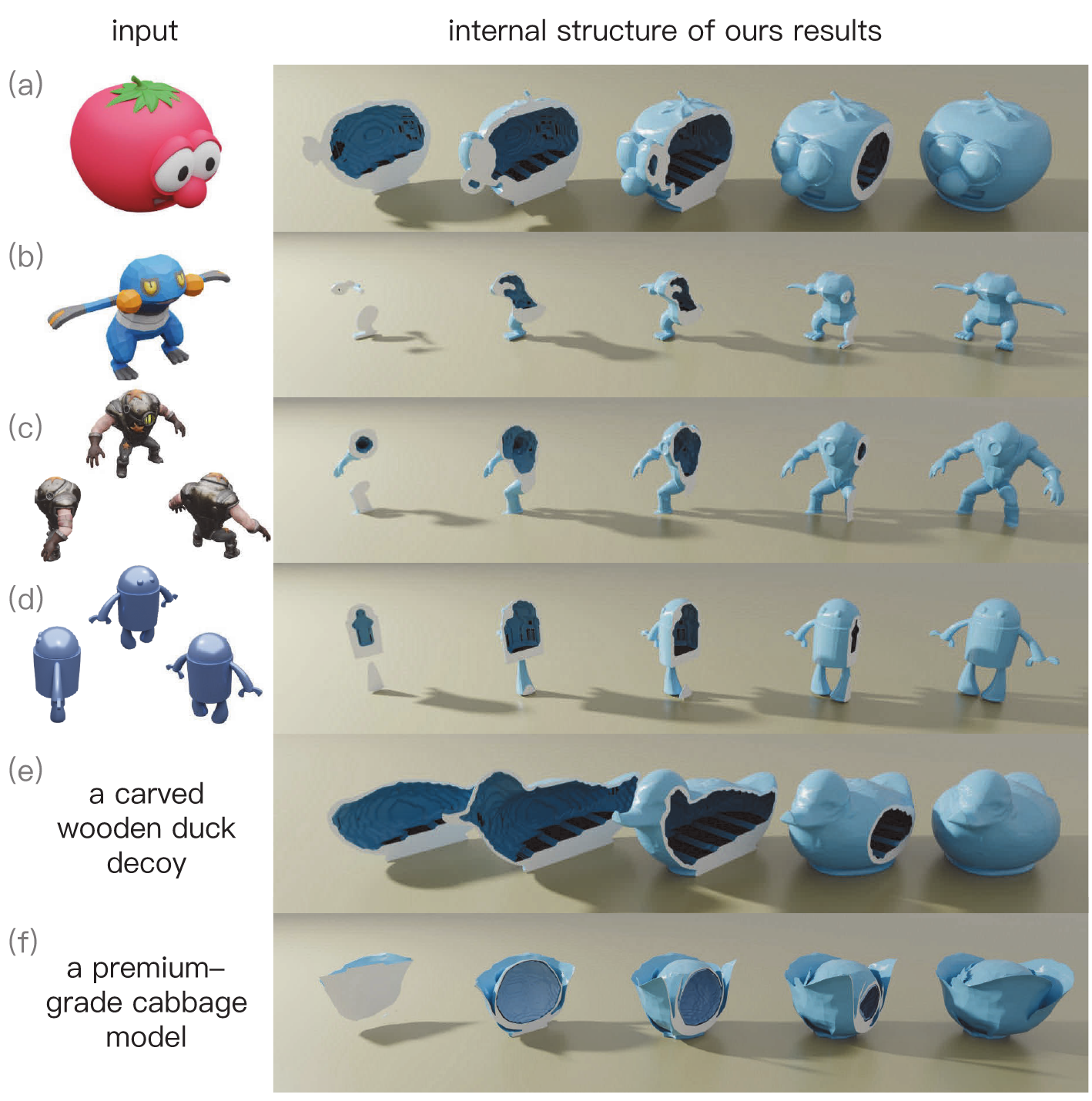}
    \caption{Self‐supporting structures generated from diverse inputs by optimizing the internal mass distribution; all examples remain upright under simulation.}
    \label{fig:inner}
\end{figure}

\begin{figure*}[ht]
    \centering
\includegraphics[width=1.0\linewidth]{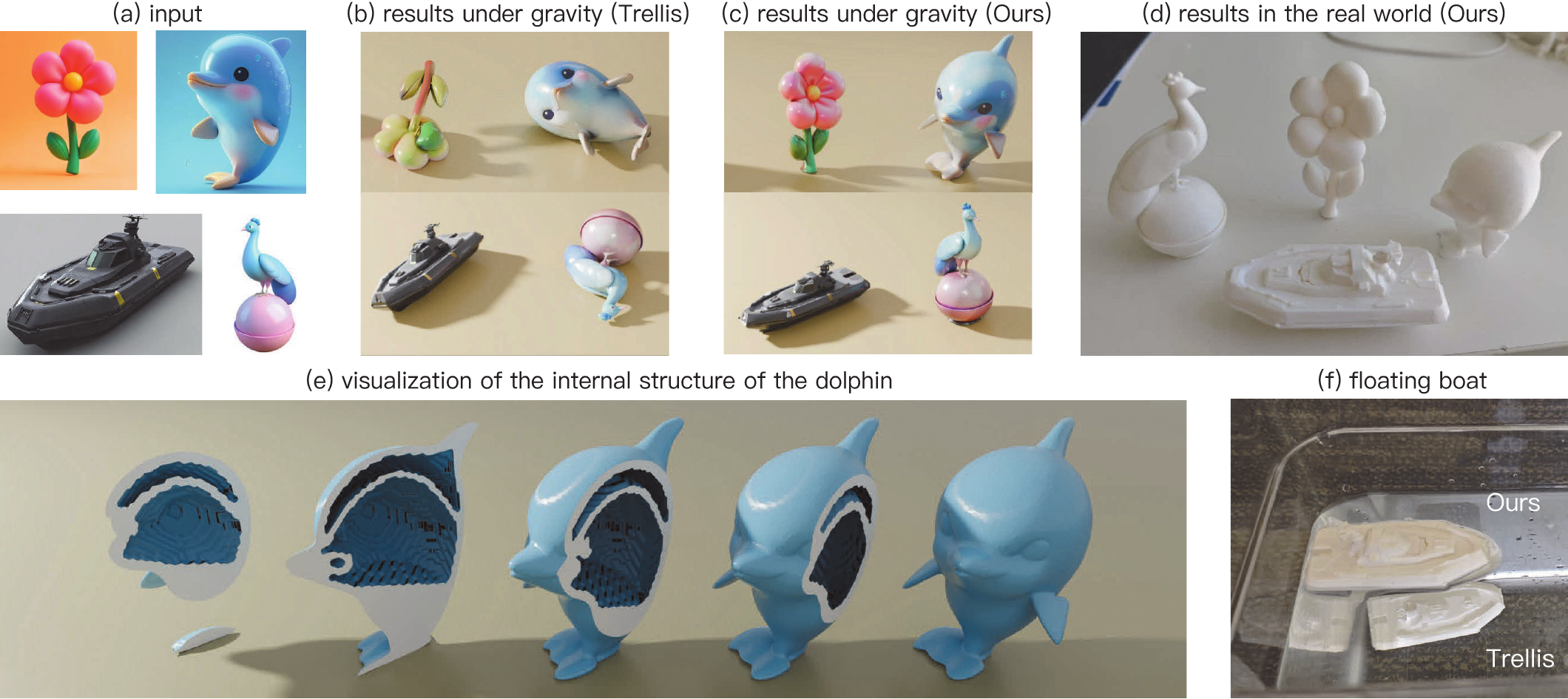}
    \caption{Our method produces self‐supporting, material‐efficient hollow structures.}
    \label{fig:teaser}
\end{figure*}
\begin{figure}
    \centering
\includegraphics[width=1.0\linewidth]{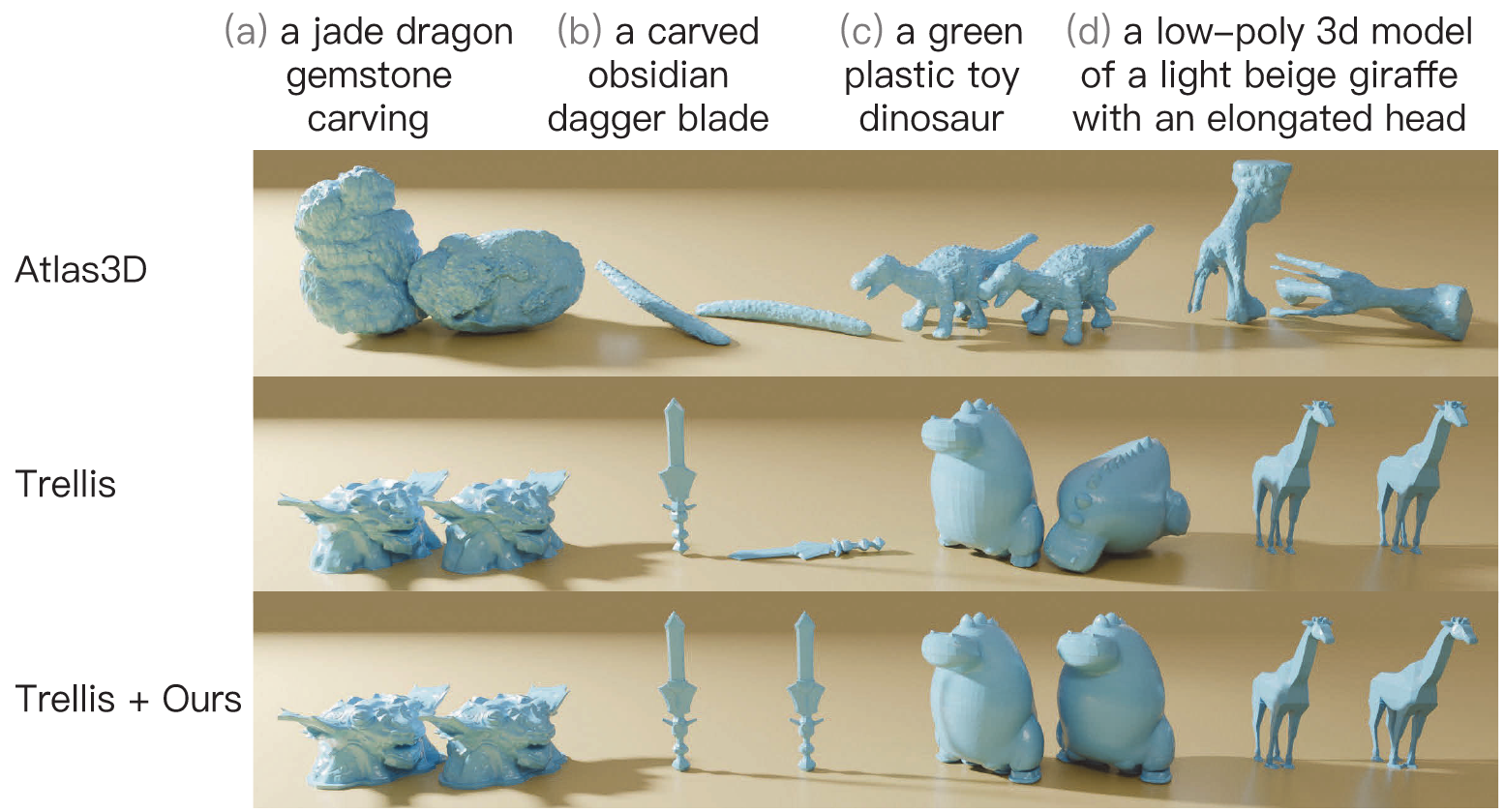}
    \caption{
Text-to-3D generation results (each pair: left: without gravity; right: under gravity). 
}\label{fig:t-to-3d}
\end{figure}
\begin{figure*}[tbp]
    \centering
    \includegraphics[width=1.0\linewidth]{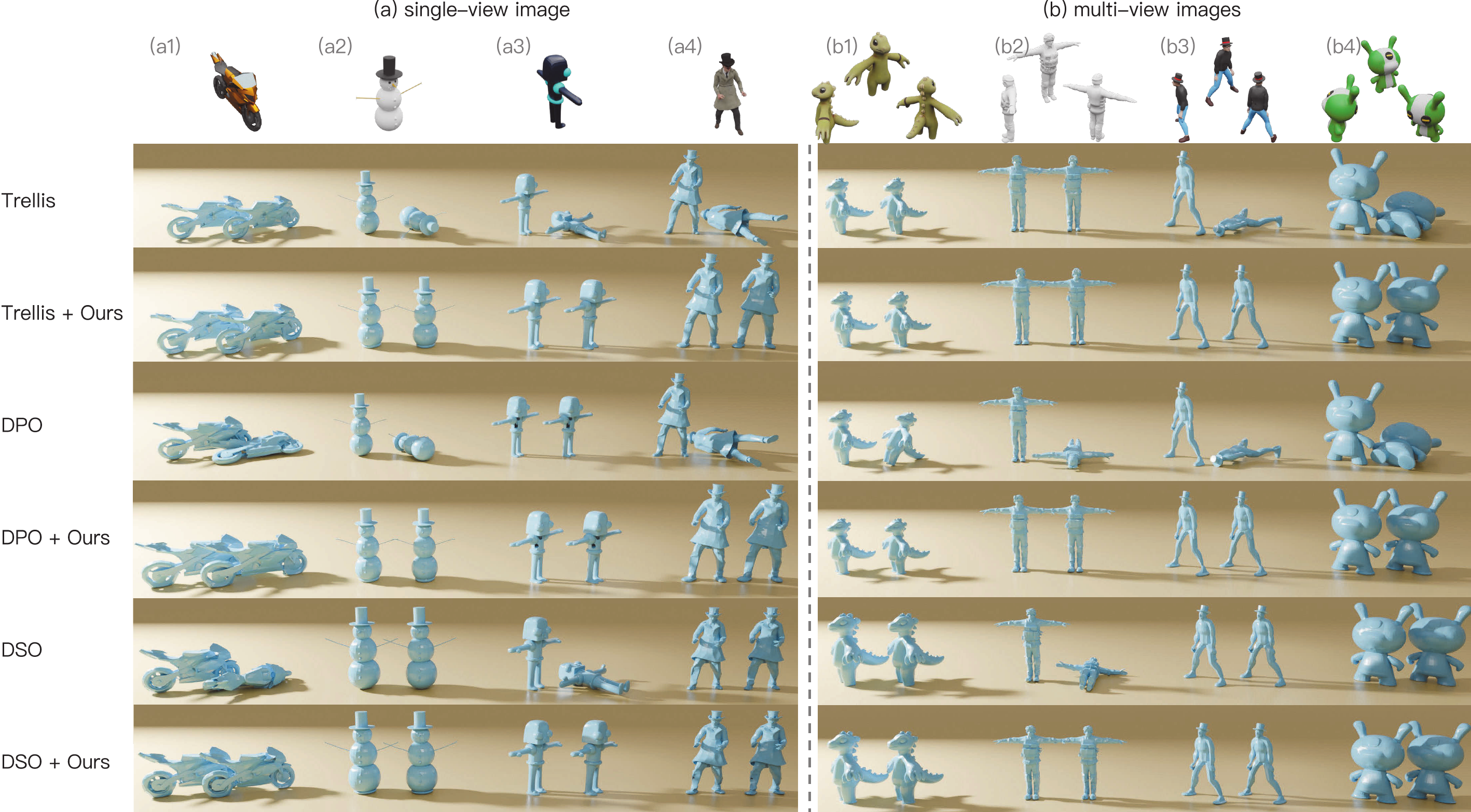}
    \caption{
Image-to-3D generation results (each pair: left: without gravity; right: under gravity). }
    \label{fig:i-to-3d}
\end{figure*}

\begin{table}[t]
\setlength{\tabcolsep}{1mm}
\centering
\begin{tabular}{lcccc}
\toprule

& Atlas3D & Trellis & DSO/DPO & Ours \\  
\midrule
%physically constrained & \CheckmarkBold &\XSolidBrush &\CheckmarkBold&\CheckmarkBold\\
physically constrained &  {\CheckmarkBold} & {\XSolidBrush} & {\CheckmarkBold}& {\CheckmarkBold}\\

%material efficient & \XSolidBrush & \XSolidBrush& \XSolidBrush& 
%\CheckmarkBold\\
material efficient &  {\XSolidBrush} &  {\XSolidBrush}&  {\XSolidBrush}& 
 {\CheckmarkBold}\\

%efficient generation &\XSolidBrush&\CheckmarkBold&\CheckmarkBold&\CheckmarkBold\\
efficient generation & {\XSolidBrush}& {\CheckmarkBold}& {\CheckmarkBold}& {\CheckmarkBold}\\

%without finetuning &\CheckmarkBold&\CheckmarkBold&\XSolidBrush&\CheckmarkBold\\
without finetuning & {\CheckmarkBold}& {\CheckmarkBold}& {\XSolidBrush}& {\CheckmarkBold}\\

%high-quality surface &\XSolidBrush&\CheckmarkBold&\CheckmarkBold&\CheckmarkBold\\
high-quality surface & {\XSolidBrush}& {\CheckmarkBold}& {\CheckmarkBold}& {\CheckmarkBold}\\
\bottomrule
\end{tabular}
\caption{Comparison of baselines and our method across key criteria.}
\label{tab:baseline-compare}
\end{table}

\section{Experiments}
In this section, we present our experimental results, including baseline comparisons, ablation study and real-world 3D-printing experiments.

\subsection{Experimental Settings}
We present the experimental settings, covering implementation, baselines, dataset details, and metrics.

\textit{Implementation.} We performed our model on a NVIDIA H100 GPU on text-to-3D and image-to-3D (single-view and multi-view) tasks. Consistent with Trellis, we set $N=64$. For hyperparameters, we used $\lambda_{\text{mass}}=100$ and $\epsilon = 2$ found on a small validation set (analysis provided in the appendix) and 42 as the global random seed. The optimization employed the Adam optimizer for 2000 steps with a learning rate of 1e$^{-2}$. All the objects were scaled to the range of $[-0.5, 0.5]^3$ for comparison. 

\textit{Baselines.}
We compare our method against SOTA baselines, including both generative models and physically-aware optimization:
(1)~\textbf{{Trellis}}~\cite{trellis} is a leading 3D generative model that produces high-quality solid geometry from text or images.  
(2)~\textbf{Atlas3D}~\cite{atlas3d} optimizes a differentiable simulation objective to encourage self-supporting geometry during synthesis on the text-to-3D task.
(3)~\textbf{DPO}~\cite{wallace2024diffusion} improves physical plausibility by learning from paired supervision~(i.e., plausible vs. implausible shapes).  
(4)~\textbf{DSO}~\cite{dso} avoids paired data by optimizing generation through direct physical reward signals derived from differentiable simulations.
We compare our method with the baselines in Table~\ref{tab:baseline-compare}.
For text-to-3D , we compare with \textbf{Trellis} and \textbf{Atlas3D}. For image-to-3D (including single- and multi-view), we include \textbf{Trellis}, \textbf{DPO}, and \textbf{DSO}.
Our method is fully compatible with Trellis‐based models. Accordingly, we also evaluate its integration with DPO and DSO.

\textit{Dataset.}
For the text-to-3D testset, we assembled 150 concise prompts generated by GPT, each describing a target 3D object (e.g., “a detailed dragon fruit model”). 
For the image-to-3D testset, we collected rendering images of unsupported, stability-challenging 3D models, from a carefully selected subset of the Objaverse‑xl~\cite{objaverse} dataset. 150 objects were collected for rendering, spanning characters, animals, and various unsupported decorative objects. For single-view image input, only a frontal rendering was used. For multi-view image inputs, three renderings from distinct viewpoints were used. 
\begin{table}[t]
\centering
\setlength{\tabcolsep}{1mm}
{
\begin{tabular}{clcccc}
% \small 
\toprule
\textbf{}& & \textbf{Mas}$\downarrow$& \textbf{Stable}$\uparrow$ & \textbf{Rot} $\downarrow$ & \textbf{CLIP}$\uparrow$\\
\midrule
\multirow{2}{*}{text}& Trellis & 0.214 & 84\%& $14.39^\circ$ &   24.4 \\
 &\enspace+Ours  & \textbf{ 0.121}~(-43\%) & \bf 92.7\% & \bf 5.48$^\circ$ & \bf 24.4 \\
 \midrule

&Trellis & 0.046 & 68.0\%&  $32.23^\circ$ & 84.7\\

single&\enspace +Ours & \textbf{0.036}~(-22\%) & 83.3\%&  $15.80^\circ$ & 84.4 \\
view&DPO & 0.055 & 83.3\%  &  $15.41^\circ$ &  84.3\\
image&\enspace +Ours & 0.044~(-19\%) & 88.7\%  &   $10.53^\circ$ &  84.1\\
 &DSO & 0.079  &91.3\%&$8.28^\circ$ &\bf 85.0\\
 &\enspace+Ours & 0.057~(-28\%)  & \bf 94\%& \bf 5.64$^\circ$ &84.5 \\
\midrule

&Trellis& 0.047 & 71.3\% &  26.49$^\circ$ &  84.6\\
multi &\enspace+Ours & \textbf{0.035}~(-26\%) & 87.3\%&  11.59$^\circ$ & 84.6 \\
view&DPO & 0.052 &  83.3\% & 14.95$^\circ$ & 84.6\\
images&\enspace+Ours & 0.039~(-24\%) & 90.7\%  &   8.19$^\circ$ &  84.4 \\
&DSO & 0.078 & 89.3\% &  10.61$^\circ$ &  \bf 84.6 \\
 &\enspace+Ours & 0.055~(-29\%)  &\bf 93.3\% &\bf 6.60$^\circ$ &84.7 \\
\bottomrule
\end{tabular}}
\caption{Quantitative evaluation of material efficiency, physical stability, and semantic consistency.}
\label{tab:vol}
\end{table}
\begin{table}[tbp]
\centering
\begin{tabular}{llcc}
\toprule
\textbf{}&  & \textbf{F-score}$\uparrow$ & \textbf{CD}$\downarrow$\\
\midrule
 text& Trellis + Ours  &\bf 0.9988&\bf 0.0081\\
 \midrule
single-view &Trellis + Ours &\bf 0.9986&\bf 0.0056\\
image&DPO  &0.9517& 0.0143\\
 &DSO  & 0.7996& 0.0336\\
\midrule
 multi-view &Trellis + Ours &\bf 0.9995&\bf 0.0054\\
images&DPO &0.9719&0.0100\\
&DSO  &0.8137& 0.0319\\

\bottomrule
\end{tabular}
\caption{Geometric fidelity of different generation methods. 
}\label{tab:geo}
\end{table}

\textit{Metrics.}
We evaluate our method and baselines from the following four aspects, including material efficiency, physical stability, semantic consistency, and geometric fidelity:
\begin{figure}
    \centering
\includegraphics[width=0.8\linewidth]{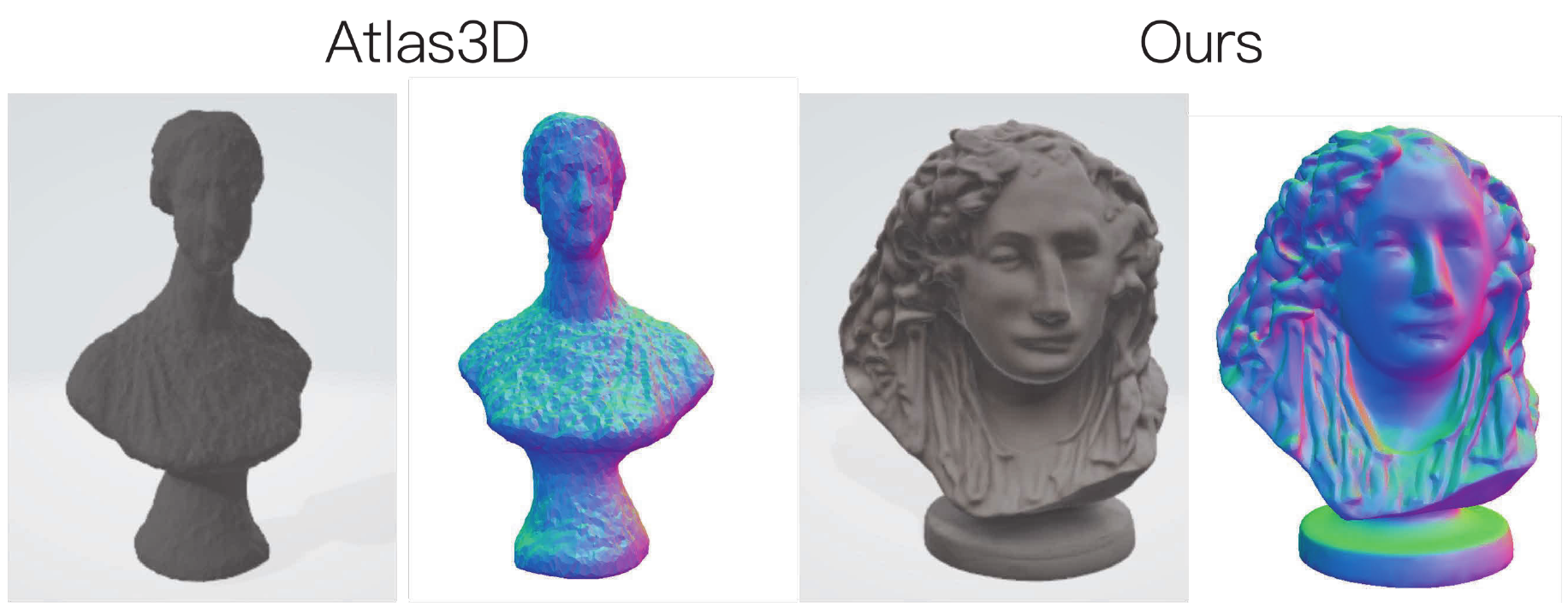}
    \caption{Surface meshes and normal maps generated by Atlas3D and our method. Atlas3D is almost unable to produce a smooth and high-quality surface mesh.}
    \label{fig:normal-map}
\end{figure}
\begin{figure}[ht]
    \centering
\includegraphics[width=0.6\linewidth]{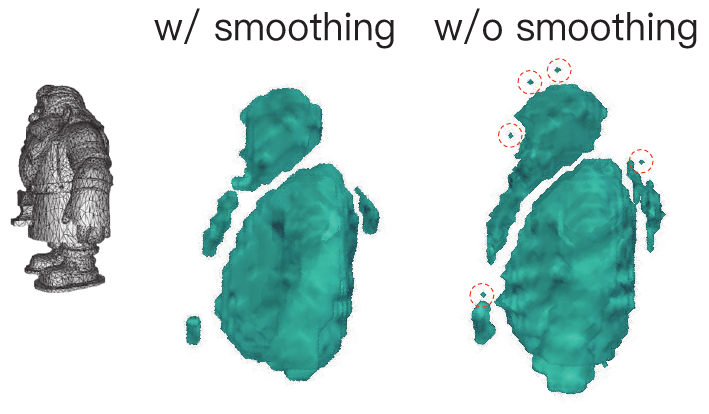}
    \caption{Ablation of density-field smoothing. Without smoothing (right), the optimized cavities contain isolated, tiny spaces~({red circle}). 
    }
    \label{fig:smoothing}
\end{figure}

    \begin{figure}[ht]
        \centering
        \includegraphics[width=\linewidth]{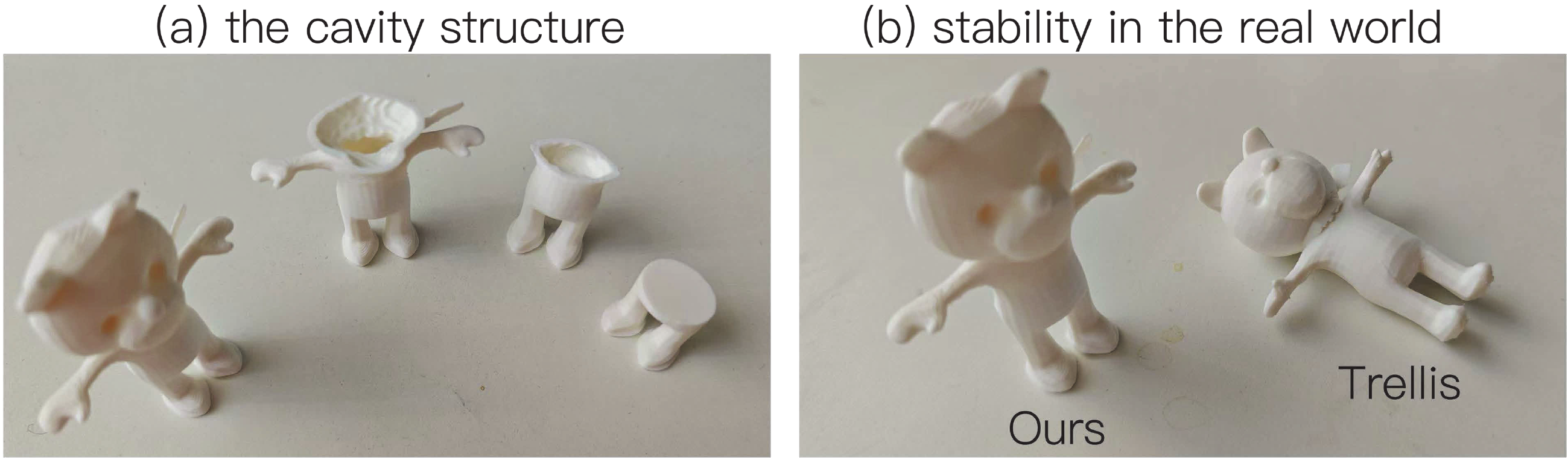}
        \caption{Real-world 3D print results: (a) the hollowed interior (b) reducing mass while maintaining upright stability under gravity, unlike solid Trellis outputs.}
         \label{fig:enter-label}
    \end{figure}

    \begin{figure}[ht]
        \centering
        \includegraphics[width=\linewidth]{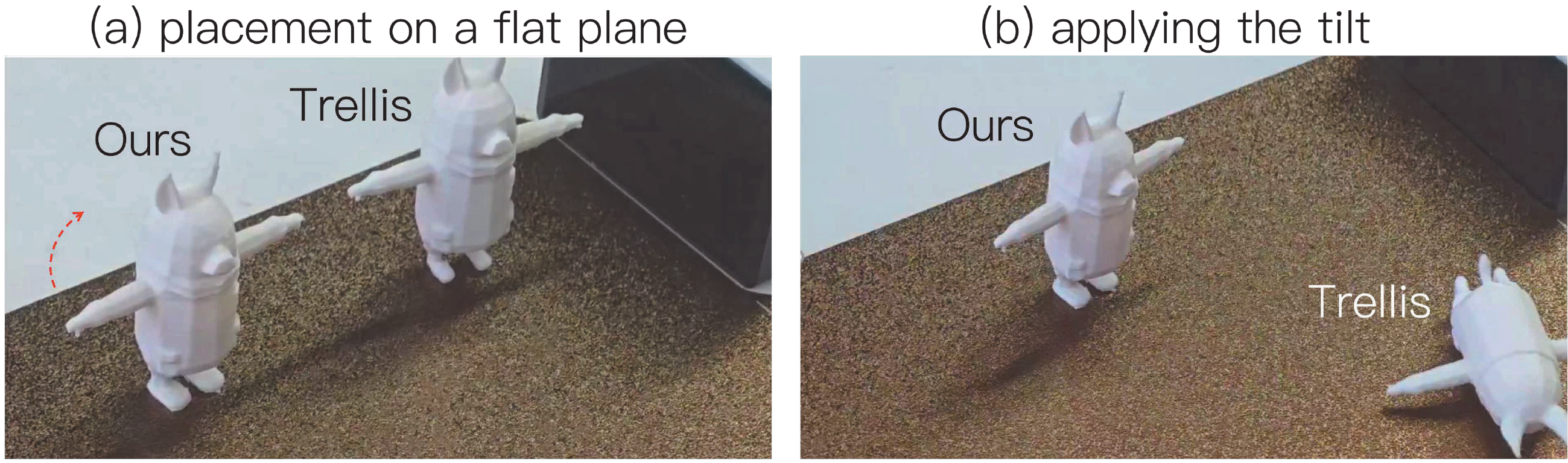}
        \caption{Stability under lean. To test robustness, we gradually lean the supporting platform.
        }
        \label{fig:rotation}
    \end{figure}
\begin{itemize}
  \item \textbf{Material Efficiency (Mas).}  
  We compute the average material mass of all generated objects~(Mas). 

  \item \textbf{Physical Stability (Stable, Rot).}  
  We run simulations in Mujoco~\cite{mujoco} to test and report the proportion of stable outcomes (Stable) and the average angular deviation from the upright pose (Rot). For hollowed structures, we recompute the inertia and center of mass for simulation.

  \item \textbf{Semantic Consistency (CLIP).}
  We measure how consistently the output matches the input by computing the CLIPScore~\cite{clipscore} between multi-view renderings of the generated object and the input~(CLIP).
  \item \textbf{Geometric Fidelity (F-score, CF).}  
  To evaluate how much the generated surface mesh deviates from the original mesh generated by Trellis, we use the Chamfer Distance (CF) and F-Score between the results of each baseline and Trellis.
\end{itemize}
\subsection{Results and Comparisons}
We first showcase representative examples, and then provide both qualitative and quantitative comparisons.

\begin{table}[tbp]
\centering
\begin{tabular}{lcccc}
\toprule
Atlas3D & Trellis & DPO&DSO & Trellis + Ours\\  
\midrule
29min & 24.3s &28.5s &28.8s&29.1s\\
\bottomrule
\end{tabular}
\caption{Wall-clock time of generating a 3D object.}\label{tab:time}
\end{table}

\begin{table}[tbp]
\centering

\begin{tabular}{lccc}
\toprule
 & \textbf{Mas}$\downarrow$& \textbf{Stable}$\uparrow$ & \textbf{Rot}$\downarrow$  \\
\midrule
Ours & 0.035 & \bf 87.3\%&  \bf 11.59$^\circ$  \\
 \midrule
w/o $\mathcal{L}_{\text{mass}}$ & 0.037 & 86.0\%&  13.17$^\circ$  \\
w/o $B$ & 0.032 & 74.7\%&  24.28$^\circ$  \\
 \midrule
 w/o $\mathcal{L}_{\text{center}}$ & 0.035 & 86.0\%&  12.26$^\circ$  \\
 w/o $\mathcal{L}_{\text{area}}$ & 0.033 & 76.7\%&  21.26$^\circ$  \\
 w/o $\mathcal{L}_{\text{z}}$ & \bf 0.030 & 85.3\%&  13.72$^\circ$  \\
\bottomrule
\end{tabular}

\caption{Ablation of different loss terms and regularization.}
\label{tab:ablation}
\end{table}

\textit{{Results.}}
Fig.~\ref{fig:inner} illustrates how our method adaptively hollows out internal regions to improve stability while preserving the visual fidelity of the input across different modalities. The hollowing pattern typically targets the upper regions of the object, where cavities can be introduced without compromising structural support. This process results in a substantial volume reduction, especially in objects with large internal mass (e.g., Fig~\ref{fig:inner}~(a,e,f)), where extensive and continuous internal voids are carved. In contrast, for objects with relatively less internal volume (e.g., Fig~\ref{fig:inner}~(b)), the optimization selectively hollows out parts of the torso while preserving denser lower-body regions to enhance standing stability.

\textit{Qualitative Comparison. }
For the text-to-3D generation~(Fig.~\ref{fig:t-to-3d}), we observe that Atlas3D struggles to generate 3D objects that reflect the input text~(e.g, dinosaur with five legs and the incomplete giraffe model). Additionally, the surface quality of Atlas3D's outputs is poor, with noticeable roughness~(Fig.~\ref{fig:normal-map}). In contrast, Trellis generates high-quality objects with smooth surfaces, but faces challenges in stability~(e.g., Fig.~\ref{fig:t-to-3d}~(b,c)). Our method ensures both high generation quality and self-supporting structures.
For the image-to-3D generation, we present results in Fig.~\ref{fig:i-to-3d} for both single-view and multi-view images. All baselines produce high-quality 3D models. Interestingly, DSO-generated objects tend to have significantly larger volumes compared to the others~(Fig.~\ref{fig:i-to-3d}~(a2, b1, b4)). In contrast, our method excels at preserving a lighter volume while still maintaining high stability, as shown by the improved stability and smaller volumes in our results.
Moreover, when integrated into Trellis, DPO, and DSO, our method consistently enhances the upright stability of their generated objects.

\textit{Quantitative Comparison.}
As shown in Table~\ref{tab:vol}, our method consistently improves both material efficiency and physical stability without degrading semantic consistency.
For text-to-3D generation, our approach reduces the average mass by 43.3\%, while increasing the stability success rate to 92.7\% and significantly lowering the average tilt angle to 5.48°.
For image-to-3D generation, our method improves over Trellis and further enhances DPO and DSO. On top of DSO, our optimization reduces mass by 28.4\% while improving stability to 94.0\% and reducing tilt by ~2.6°.
Our method introduces minimal distortion while enhancing functionality. Table~\ref{tab:geo} shows that our method~(best F-Score and CD) achieves the highest geometric fidelity of the original Trellis outputs. 
Table~\ref{tab:time} reports the wall‐clock time to generate a 3D design. Atlas3D requires nearly 29 minutes, whereas Trellis and related methods (DPO, DSO) finish in under 30 seconds. Our density‐field optimization adds only a few seconds of overhead beyond Trellis.

\subsection{Ablation Study}
We performed an ablation study under the multi-view images input setting. 

\textit{Density Smoothing.} 
As shown in Fig.~\ref{fig:smoothing}, we visualize the inner surfaces of the carved cavities. With smoothing, the hollow regions form large, continuous pockets; without smoothing, the inner surface contains small, isolated areas that are fragile and easily lost during fabrication.  

\textit{Different Physical Constraints.}  
Table~\ref{tab:ablation} shows that each physics‐based loss contributes uniquely. Removing  
\(\mathcal{L}_{\mathrm{center}}\) slightly degrades stability, and 
removing \(\mathcal{L}_{\mathrm{area}}\) causes a more pronounced drop in upright success.
Interestingly, disabling \(\mathcal{L}_{z}\) yields marginally better mass reduction but at the expense of a higher average tilt angle, revealing a trade-off between material efficiency and static stability.

\textit{Regularization.} 	
As shown in Table~\ref{tab:ablation}, removing \(\mathcal{L}_{\mathrm{mass}}\) slightly increases mass and reduces stability, confirming its role in promoting lightweight, balanced designs. In contrast, removing \(B\) drastically degrades stability, highlighting the importance of reinforcing ground contact regions.

\subsection{Real World 3D-Printing}
We validated the manufacturability by printing several optimized models on a standard FDM printer~(Fig.~\ref{fig:teaser}~(d), Fig.~
\ref{fig:enter-label}, Fig.~\ref{fig:rotation}). As shown in Fig.~\ref{fig:enter-label}, our method automatically carved internal cavities and reduced mass, yet the printed prototype remained upright. In contrast, the solid Trellis model failed. 
To assess robustness, we placed the 3D-printings of both Trellis‐generated and our generated structures upright on a flat platform and increased the lean angle (Fig.~\ref{fig:rotation}). Trellis model began to fail at relatively small inclinations, while our hollow design remained stable. 

\section{Discussion and Conclusion}
We have presented DensiCrafter, a density‐field optimization method that generates lightweight, self‐supporting 3D structures by carving continuous internal cavities under fully differentiable, simulation‐free physical constraints. Extensive evaluations and real‐world 3D-printing experiments confirm that our method delivers significant material savings, enhanced upright stability, and reliable manufacturability without compromising visual fidelity and computational efficiency. Failure cases and corresponding analyses are provided in the appendix. By integrating the generative model with practical stability and fabrication requirements, DensiCrafter paves the way for deployable, physics‐aware 3D content creation across text‐to-3D and image‐to-3D generation pipelines.   
Relative to Spin‑it Faster~\cite{hafner2024spin}, which segments density fields via quadric surface fitting, DensiCrafter directly optimizes the continuous voxel field. This direct approach more naturally leverages Trellis’s structure latent representation.

While DensiCrafter produces lightweight, self‐supporting hollow structures effectively, it assumes all objects behave as rigid bodies and does not model internal stress distributions or deformations under load. Extending our method to handle elastic or compliant materials would require mechanics-aware constraints or differentiable finite element methods. DensiCrafter may also inherit distortions from Trellis, and enhancing base model quality is valuable future work. Another promising future direction is to learn the prior of mass distribution from large-scale data, enabling the development of world models~\cite{ding2024understanding} that can reason about the real-world physical laws.

\section{Acknowledgments}
Nan Cao is the corresponding author. This work was sponsored by Natural Science Foundation of Shanghai (25ZR1401336).

\bibliography{aaai2026}

\newpage

\newpage

\appendix 
\section{Appendix}

\subsection{Proof of Centroid in Convex Hull}

We provide a detailed proof of the claim that the centroid of the support region $S$ always lies within its convex hull.

\subsubsection{Claim.}
Let $S \subset \mathbb{R}^2$ denote the bottom contact region on the ground plane, defined as the set of points where $\tilde{\rho}(\mathbf{x}) > 0.5$ at $z = z_{\min}$. 
Assume that $S$ is measurable, bounded, and has finite nonzero area $|S| = \int_S {\rm d}A < \infty$. 
Let $c$ denote the centroid of $S$. 
Then the centroid lies within the convex hull of $S$, i.e.,
$c \in \mathrm{conv}(S)$.

\subsubsection{Proof.}
The centroid of $S$ is defined as
\begin{equation}
c = \frac{1}{|S|} \int_S \mathbf{x}\, {\rm d}A
\label{eq:centroid}
\end{equation}
where ${\rm d}A$ denotes the differential area element on the ground plane.
The convex hull of $S$ is the smallest convex set containing $S$, given by
\begin{equation}
\mathrm{conv}(S) = 
\left\{
\sum_{i=1}^k \lambda_i \mathbf{x}_i \;\middle|\;
\mathbf{x}_i \in S,\,
\lambda_i \ge 0,\,
\sum_{i=1}^k \lambda_i = 1,\,
k \in \mathbb{N}
\right\}
\end{equation}

To show that $c \in \mathrm{conv}(S)$, we approximate the integral in~\eqref{eq:centroid} by a sequence of weighted Riemann sums. 
Partition $S$ into $n$ measurable subregions $S_i^{(n)}$ with respective areas $A_i^{(n)} \ge 0$, and select a representative point $\mathbf{x}_i^{(n)} \in S_i^{(n)}$. 
Then the Riemann sum approximation of the centroid is
\begin{equation}
c_n = \frac{1}{|S|} \sum_{i=1}^n A_i^{(n)}\, \mathbf{x}_i^{(n)}
     = \sum_{i=1}^n w_i^{(n)}\, \mathbf{x}_i^{(n)},
\end{equation}
where $w_i^{(n)} = A_i^{(n)}/|S| \ge 0$ and $\sum_{i=1}^n w_i^{(n)} = 1$. 
Each $c_n$ is therefore a convex combination of points in $S$, implying $c_n \in \mathrm{conv}(S)$ for every finite $n$.
As the partition is refined, the sequence $\{c_n\}$ converges to the integral limit $c$. 
Since $\mathrm{conv}(S)$ is closed for any bounded $S \subset \mathbb{R}^2$, the limit point $c$ also belongs to $\mathrm{conv}(S)$. 
Hence, $c \in \mathrm{conv}(S)$.

\subsection{Compatibility}

While implemented with voxel-based representations from Trellis, our method is representation-agnostic. The formulation extends naturally to other volumetric representations including signed distance fields (SDFs) and neural implicit fields, enabling integration with diverse 3D generative pipelines without modifying the stability constraints or optimization objectives.

\begin{figure}[ht]
\centering
\includegraphics[width=1.0\linewidth]{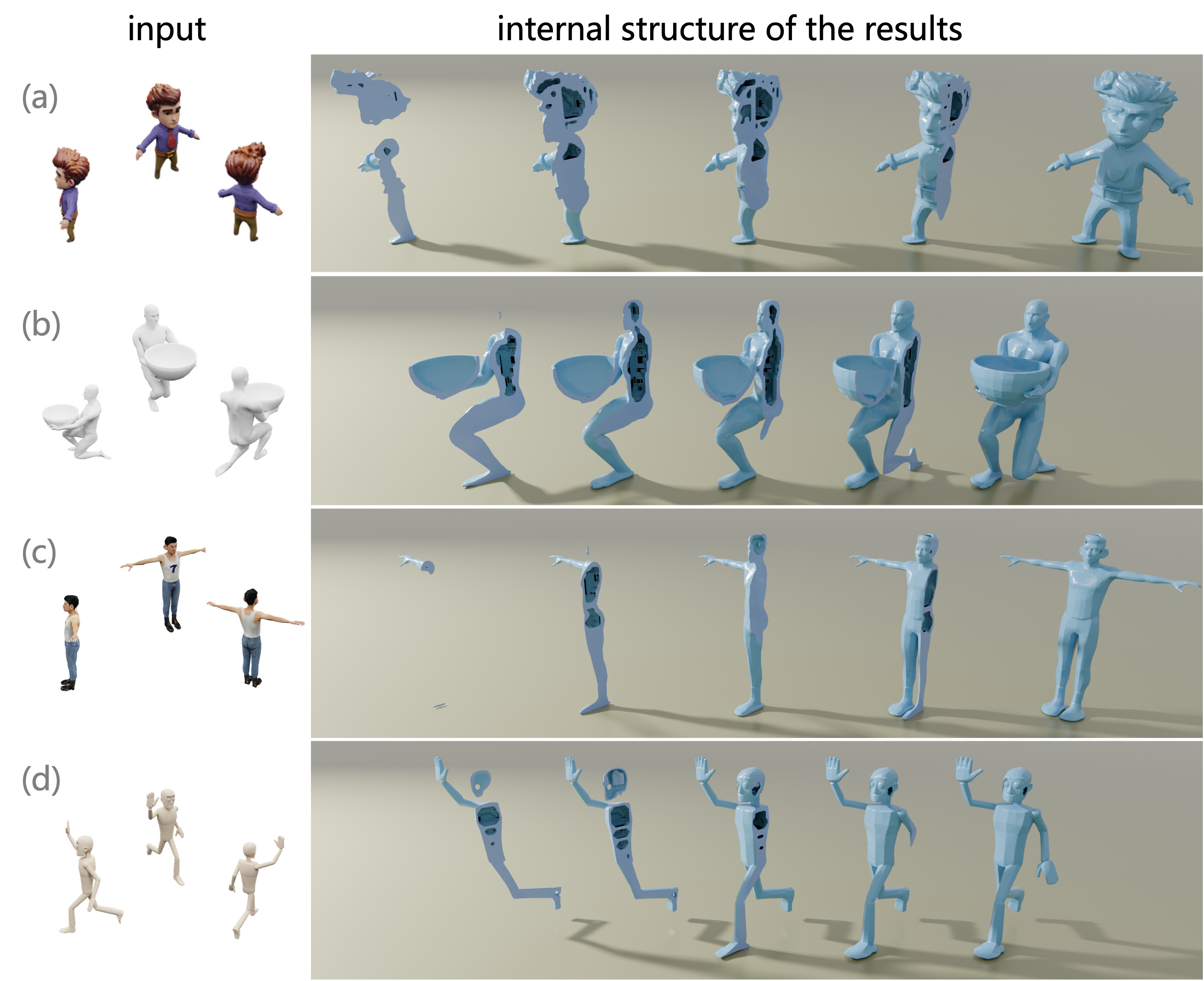}
\caption{Failure cases.}
\label{fig:failures}
\end{figure}

\subsection{Failure Analysis}

We provide additional failure cases and analysis. As shown in Figure~\ref{fig:failures}, our method encounters challenges with slender structures and inherently unstable poses. For instance, running humanoid figures with minimal ground contact area present fundamental stability limitations that cannot be resolved through internal hollowing alone. These cases highlight the inherent physical constraints of static stability that cannot be overcome through mass redistribution when the base support area is insufficient.

\subsection{Hyperparameter Analysis}

We analyze the effect of the hyperparameters (i.e., the mass regularization weight $\lambda_{\rm mass}$ and the basal layer thickness $\epsilon$) on geometric quality and structural stability. A small-scale experiment was conducted on the validation set to determine reasonable values for these parameters.
\begin{table}[htbp]
\centering
\begin{tabular}{lcccc}
\toprule
 $\lambda_{\rm mass}$& \textbf{Mass}$\downarrow$ & \textbf{Stable}$\uparrow$ & \textbf{Rot}$\downarrow$ $\uparrow$ & \textbf{CD}$\downarrow$\\
\midrule
$0$ & 0.038 & 92\% & 8.46$^\circ$  & 0.0058\\
$10$ & 0.038 & 92\% & 8.48$^\circ$  &0.0057\\
$100$ & 0.035 & 92\% & 8.44$^\circ$ &0.0058\\
$1000$ & 0.031 & 90\% & 13.86$^\circ$ &0.0057\\
\bottomrule
\end{tabular}
\caption{Effect of $\lambda_{\rm mass}$ on material usage and stability. Higher values reduce mass but compromise stability when excessive.}
\label{tab:lambda-mass}
\end{table}
\subsubsection{Effect of $\lambda_{\rm mass}$.}

We evaluate $\lambda_{\rm mass}$ values under multi-view settings (Table~\ref{tab:lambda-mass}). Increasing $\lambda_{\rm mass}$ reduces material usage (Mass), but excessive values ($\lambda_{\rm mass}=1000$) compromise stability, indicated by increased rotational deviation (Rot). We select $\lambda_{\rm mass}=100$ as it achieves optimal balance between mass reduction and stability preservation.

\begin{table}[ht]
\centering
\begin{tabular}{lcccc}
\toprule
 $\epsilon$& \textbf{Mass}$\downarrow$ & \textbf{Stable}$\uparrow$ & \textbf{Rot}$\downarrow$ & \textbf{CD}$\downarrow$\\
\midrule
$1$ & 0.034 & 72\% & 28.40$^\circ$ & 0.0051\\
$2$ & 0.036 & 90\% & 10.57$^\circ$ & 0.0055\\
$3$ & 0.041 & 94\% & 6.79$^\circ$ & 0.0066\\
\bottomrule
\end{tabular}
\caption{Effect of basal layer thickness $\epsilon$ on stability and mass efficiency.}
\label{tab:epsilon}
\end{table}

\subsubsection{Effect of $\epsilon$.}

The basal layer thickness $\epsilon$ significantly impacts stability (Table~\ref{tab:epsilon}). While $\epsilon=1$ achieves the lowest mass, it provides insufficient support (72\% stability). We choose $\epsilon=2$ as it maintains high stability (90\%) with reasonable mass reduction and fairly good surface preservation.

\end{document}